\documentclass[letterpaper, 10 pt, conference]{ieeeconf}  

\IEEEoverridecommandlockouts                              

\usepackage{graphics} 
\usepackage{epsfig} 
\usepackage{times} 
\usepackage{amsmath} 
\usepackage{amssymb}  
\usepackage{multicol}
\usepackage[bookmarks=true]{hyperref}
\usepackage{array}
\usepackage{hyperref}
\usepackage{amsmath}
\usepackage{cleveref}
\usepackage{paralist}
\usepackage{pdfpages}
\let\labelindent\relax
\usepackage{enumitem}
\usepackage{wrapfig}
\usepackage{booktabs} 
\usepackage{multirow}
\usepackage{hyperref}
\usepackage{url}
\usepackage{amssymb}
\usepackage{graphicx}
\usepackage{CJKutf8}
\usepackage{algorithm}
\usepackage[noend]{algorithmic}
\usepackage{colortbl}
\usepackage{threeparttable}
\usepackage{mdframed}
\usepackage{tcolorbox}
\tcbuselibrary{minted}
\usepackage{blindtext}
\usepackage{multicol}
\usepackage{caption, subcaption}
\usepackage[utf8]{inputenc}
\usepackage{tcolorbox}
\usepackage{listings}
\usepackage{xcolor}
\usepackage{tabularx}
\usepackage{cite}
\usepackage{stfloats}
\definecolor{customblue}{HTML}{ccf2f5}
\usepackage{soul, color, xcolor}

\title{\LARGE \bf ShapeForce: Low-Cost Soft Robotic Wrist for \\Contact-Rich Manipulation}

\author{Jinxuan Zhu$^{*}$$^{1,2}$, Zihao Yan$^{*}$$^{1,2}$, Yangyu Xiao$^{1,2}$, \\Jingxiang Guo$^{1}$, Chenrui Tie$^{1}$, Xinyi Cao$^{1}$, Yuhang Zheng$^{1}$ and Lin Shao$^{\dagger}$$^{1, 2}$
\thanks{*~Equal contribution.}%
\thanks{$\dagger$~Corresponding author: Lin Shao (\texttt{linshao@nus.edu.sg}).}%
\thanks{$^{1}$ School of Computing, National University of Singapore, Singapore.}%
\thanks{$^{2}$ RoboScience, Shenzhen, China.}%
}

\begin{document}

\maketitle
\thispagestyle{empty}
\pagestyle{empty}


\begin{abstract}
Contact feedback is essential for contact-rich robotic manipulation, as it allows the robot to detect subtle interaction changes and adjust its actions accordingly. Six-axis force-torque sensors are commonly used to obtain contact feedback, but their high cost and fragility have discouraged many researchers from adopting them in contact-rich tasks. To offer a more cost-efficient and easy-accessible source of contact feedback, we present \textbf{ShapeForce}, a low-cost, plug-and-play soft wrist that provides force-like signals for contact-rich robotic manipulation. Inspired by how humans rely on relative force changes in contact rather than precise force magnitudes, ShapeForce converts external force and torque into measurable deformations of its compliant core, which are then estimated via marker-based pose tracking and converted into force-like signals. Our design eliminates the need for calibration or specialized electronics to obtain exact values, and instead focuses on capturing force and torque changes sufficient for enabling contact-rich manipulation. Extensive experiments across diverse contact-rich tasks and manipulation policies demonstrate that ShapeForce delivers performance comparable to six-axis force-torque sensors at an extremely low cost. More details of this project can be found at our project page: https://shapeforce.github.io/.

\end{abstract}

\section{Introduction}

Contact-rich manipulation is crucial for moving robots beyond simple pick-and-place capabilities toward performing real-world, skillful tasks. Tasks such as peg insertion, assembly, and screw tightening often require detecting subtle interaction transitions, such as whether a peg has aligned with a hole or whether a screw has been fully tightened. In these scenarios, visual perception alone is insufficient for monitoring task progress, verifying success, or perceiving occluded events. In contrast, contact feedback provides a complementary sensing modality, allowing the robot to infer interaction states and decide when to adapt its action policies.

Traditionally, contact sensing has been achieved using dedicated six-axis force-torque (FT) sensors \cite{chao1997shape,cao2021six,lu2006force}. These devices rely on rigid mechanical structures with strain gauges or piezoelectric elements to deliver high-precision force and torque signals. However, they are typically priced in the thousands of U.S. dollars and are prone to damage under intensive contact forces, further increasing maintenance and replacement costs. These limitations highlight the need for a cost-effective sensing solution that can provide reliable contact feedback and support contact-rich manipulation tasks.

\begin{figure}[t]
\centering
\includegraphics[width=\linewidth]{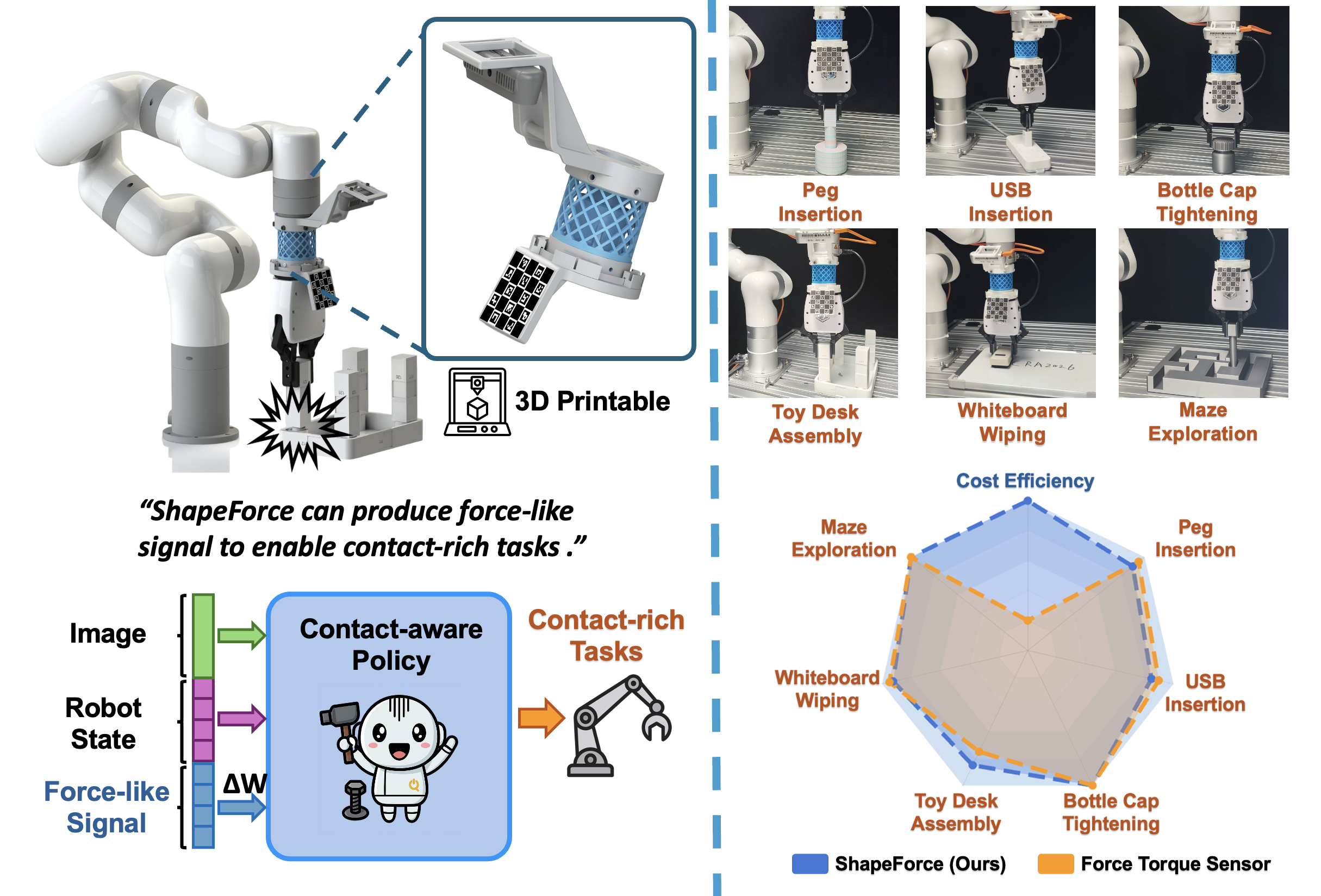}
\caption{
\textbf{ShapeForce}: a low-cost, plug-and-play sensing system that provides force-like signals, enabling contact-rich tasks with comparable performance of force–torque sensors.}

\label{fig:teaser}
\end{figure}

Several prior works~\cite{wang2021gelsight, wang2022tacto, pattabiraman2025eflesh, she2021cable, neto2021soft, Dong2025TeachingOral, Guo2024ProprioceptiveState, Liu20242024ProprioceptiveLearning, Li2025ActiveSPN} have explored alternative approaches to low-cost contact-modality sensing. However, they remain difficult to widely reproduce and apply in contact-rich tasks for two main reasons: (1) they require specialized electrical components~\cite{wang2021gelsight, wang2022tacto, pattabiraman2025eflesh, she2021cable, neto2021soft}; and (2) they rely on force–torque sensors for data collection and calibration~\cite{Dong2025TeachingOral, Guo2024ProprioceptiveState, Liu20242024ProprioceptiveLearning, Li2025ActiveSPN}. In both cases, these requirements mainly serve to obtain precise ground-truth force and torque measurements, which creates a clear conflict between the need for accurate sensing and the goal of low-cost and widely reproducible fabrication.

Taking inspiration from how humans perform contact-rich tasks, we note that when inserting a key into a lock or tightening a bottle cap, humans rarely access the exact numerical value of the applied force and torque. Instead, we rely on perceiving relative changes in force and torque to detect key interaction events and adjust our actions at the right moment. This observation suggests that precise ground-truth measurement is not strictly necessary for contact-rich manipulation, as long as the system can capture reliable trends and relative changes in interaction forces. This insight provides a way to resolve the conflict between accuracy and simplicity: by focusing on relative force and torque changes rather than exact force–torque values, we can remove the dependence on expensive force–torque sensors for calibration and specialized electronics for measurement, enabling a design that is low cost, easy to fabricate, and widely reproducible, while still delivering reliable contact feedback.

Guided by this idea, we present ShapeForce, a low-cost, plug-and-play soft wrist that converts external forces and torques into measurable deformations of its compliant core. These deformations are subsequently captured by a wrist-mounted RGB camera and transformed into six-dimensional force-like signals via marker-based pose tracking. Our design removes the need for expensive force-torque sensors or specialized electronics, instead relying only on a 3D-printed material and a wrist-mounted RGB camera commonly available in fine-grained robotic manipulation setups. This makes ShapeForce extremely easy to fabricate, deploy, and integrate into existing systems, while still preserving the contact information necessary to detect contact events.

We evaluate ShapeForce on a diverse suite of contact-rich tasks, including peg insertion, USB insertion, bottle cap tightening, toy desk assembly, whiteboard wiping, and maze exploration. We integrate it with classical algorithms for contact-rich tasks, including search routines (e.g., hole-finding) and force-control strategies, collectively denoted as search-and-control policy, as well as learning-based policy.
Across all tasks and policy settings, ShapeForce consistently improves task success compared to baselines without contact feedback and matches the performance of systems using force-torque sensors, while offering significantly lower fabrication cost and a simple, reproducible design. 

Our key contributions are as follows:

\begin{itemize}
\item We present \textbf{ShapeForce}, a low-cost, plug-and-play soft robotic wrist that captures its proprioceptive deformation and converts it into force-like signals to support contact-rich manipulation tasks.

\item Inspired by human perception, we resolve the accuracy-simplicity trade-off by providing a task-effective force representation based on relative changes, further reducing the reproduction and deployment burden.

\item We integrate our sensing system with both classical search-and-control and learning-based policies, and show experimentally that it consistently improves task success across diverse contact-rich scenarios, demonstrating the effectiveness of our force-like signals.

\end{itemize}
\section{Related Work}

\subsection{Mechanical Design of Robotic Wrists}

The human wrist enables highly dexterous manipulation, inspiring numerous robotic wrist designs. Prior works \cite{peticco2025dexwrist,negrello2019compact,sun2024compact,lemerle2021configurable,taheri2023design} have explored compact and variable-stiffness architectures to expand manipulator workspace, improve flexibility, and facilitate manipulation in constrained environments. While these actively actuated designs provide precise and stable control, they often increase system complexity, reduce robustness under collision, and raise manufacturing costs.

To mitigate these drawbacks, several studies \cite{jeong2025biflex,kurumaya2018modular,von2020compact,zhang2020stiffness} have investigated passively compliant wrists to reduce cost while enhancing flexibility. The main challenge of passive compliance lies in balancing flexibility with positional precision. Recent works, such as BiFlex \cite{jeong2025biflex}, leverage controlled buckling \cite{budiansky1974theory} to achieve bimodal stiffness, inspiring us to tune design parameters to balance compliance and stability.

Unlike these prior efforts, our approach not only exploits passive compliance but also leverages deformation as a sensing modality to estimate external forces and torques, enabling a contact-aware policy for contact-rich manipulation.

\subsection{Vision-based Deformation Perception}

Vision-based Deformation Perception (VBDeformP) is a perception paradigm that leverages visual sensing to estimate the deformation of objects or structures, which are subsequently used for force sensing\cite{wan2022visual,Dong2025TeachingOral,Li2025ActiveSPN,zhu2025forces}, reconstruction\cite{Guo2024ReconstructingSoft,Liu20242024ProprioceptiveLearning}, and tactile perception\cite{Guo2024ProprioceptiveState, xu2024vision}. These deformable structures are usually made of thermoplastic polyurethane (TPU) and can be manufactured by 3D printing, providing a low-cost perceptual approach. 

A major challenge lies in representing the deformation of a soft body, which theoretically has infinite degrees of freedom. Prior works address this by using fiducial markers~\cite{Wu2024VisionBasedSRM} or segmentation models such as SAM~\cite{zhu2025forces} to extract features, followed by regression models trained with paired force-torque data for deformation-to-force mapping. In contrast, we constrain deformation using a hard-soft-hard sandwich-like structure and encode it as the relative pose of two rigid ends, which is subsequently turned into single-marker pose tracking. The structural design enables a easier deformation estimation and benefits the deformation as representation.

\subsection{Reproducible and Accessible Sensor Designs}
Several prior works propose reproducible and accessible contact-sensing solutions through different mechanisms.
eFlesh \cite{pattabiraman2025eflesh} combines parameterized cut-cell microstructures with off-the-shelf Hall effect sensors to build a highly customizable magnetic tactile sensor. Vision-based tactile sensors such as DIGIT \cite{lambeta2020digit} use a compact camera and a transparent elastomer to visually capture surface deformations. Building on this foundational idea, ReSkin \cite{piacenza2021reskin}, AnySkin \cite{bhirangi2409anyskin}, and the TacTip family \cite{ward2018tactip} have further expanded the capabilities of vision-based tactile sensing.

However, most of these methods focus on recovering absolute force values or detailed tactile signals, which typically require calibration with ground-truth data or additional specialized electronics. In contrast, our design can be fabricated with a standard laboratory 3D printer with easy-accessible materials and uses only a wrist-mounted RGB camera to perceive, a setup that is already common in robotic manipulation. This simple and reproducible design lowers the fabrication and deployment barrier and increases its potential for adoption in both research and real-world applications.

\section{Methodology}

Our methodology consists of three components. We first present the mechanical design and fabrication, focusing on how the compliant core achieves desired deformation behavior (Sec.~\ref{method: design}). We then describe capturing deformation via marker-based pose tracking and converting it into force-like signals (Sec.~\ref{method: Deformation Capture and Representation}). Finally, we show how these signals are integrated into search-and-control policies and learning-based policies to enhance contact-rich tasks (Sec.~\ref{method: policy}).

\begin{figure*}[t]
\centering
\includegraphics[width=\linewidth]{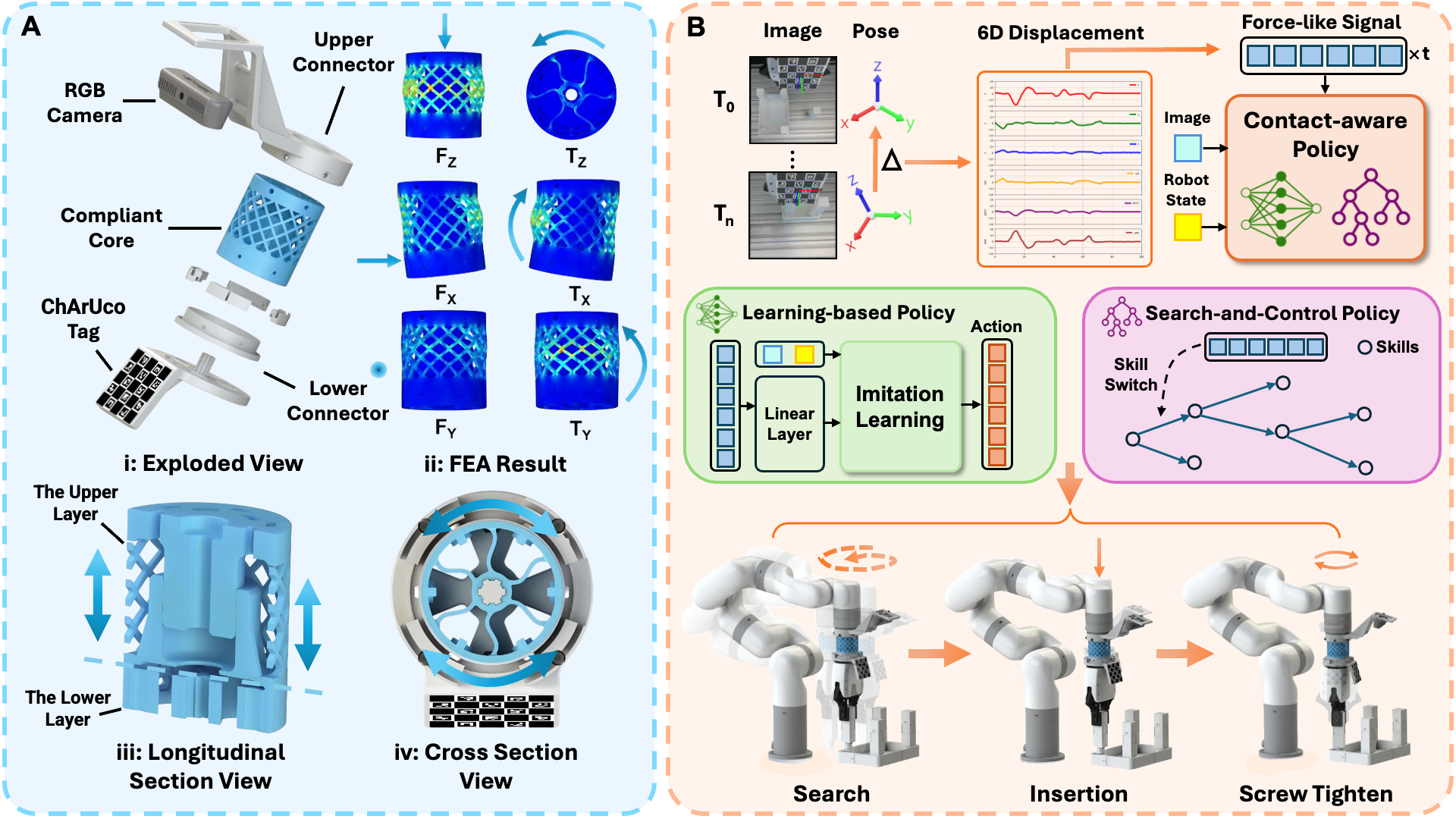}
\caption{
\textbf{Overview of ShapeForce}: (A) Mechanical design of ShapeForce, including (i) exploded view, (ii) finite element analysis (FEA) results, (iii) longitudinal section, (iv) cross section; (B) Perception-to-control pipeline. Marker-based pose tracking converts deformation into 6D displacement, producing force-like signals that inform contact-aware policies. Both search-and-control policies and imitation-learning policies are supported, enabling execution of contact-rich tasks.}
\label{fig:pipeline}
\end{figure*}

\subsection{Mechanism Design and Fabrication}
\label{method: design}
We followed two key principles for design and fabrication.
First, the sensor is designed to provide a high-quality representation of external force and torque with sufficient sensitivity, while maintaining stability and precision under dynamic motion.
Second, the system should be low-cost and simple to fabricate using easy-accessible materials, and also easy to assemble. Based on these two principles, the detailed mechanism design and fabrication process are as follows:

\subsubsection{Mechanism Design}

The mechanism design is illustrated in the exploded view in Fig.~\ref{fig:pipeline}A(i). ShapeForce consists of two main components: (1) a central compliant core (blue) that converts external forces and torques into measurable deformations, and (2) rigid connectors (white) that both link the core to the robot and constrain its deformation.

The compliant core is the centerpiece of our system, designed to translate external wrenches into measurable proprioceptive deformation. In typical contact-rich manipulation with a wrist-mounted gripper, the dominant interaction wrench components are the axial force $F_z$ (e.g., during insertion or pressing) and the torsional moment $T_z$ (e.g., during screwing or twisting). External forces and torques along other directions can often be transformed through suitable robotic arm pose adjustments into equivalent effects along $F_z$ or $T_z$ of the compliant core. Our design therefore emphasizes these components, mechanically decoupling $F_z$ and $T_z$ as illustrated in Fig.~\ref{fig:pipeline}A(iii)(iv), providing high-quality representation while allowing their sensitivities to be independently tuned to match task requirements. The remaining wrench components are tuned to maintain sufficient stiffness for stability and precise motion, while preserving passive compliance that improves robustness during contact. Building on this design principle, the compliant core is divided into two functional layers that decouple and emphasize the axial force $F_z$ and torsional moment $T_z$.

\textbf{The upper layer} adopts a hollow cylindrical shell with a lattice of square cutouts as the primary deformable structure~\cite{seharing2020review}, which provides reliable deformation while reducing weight and material usage. A centrally placed flexible sleeve (Fig.~\ref{fig:pipeline}A, iii) increases stiffness against lateral forces $F_x$ and $F_y$, while remaining compliant to $F_z$. This selective stiffening balances deflection, suppresses undesired lateral motion, and mitigates oscillations induced by gripper inertia, ultimately improving stability and positioning accuracy.

\textbf{The lower layer} consists of two concentric rings of different radii connected by wave-like support structures, as shown in Fig.~\ref{fig:pipeline}A(iv). The inner ring is mechanically coupled to the rigid base via a spline connection, while the outer ring is attached to the rigid circle frame and constrained by a circle rail. When an external torsional moment $T_z$ is applied, the inner and outer rings undergo relative rotation, producing an angular displacement that is quasi-linear to the applied torque. This configuration enables a relatively stable and measurable torsional deformation.

\subsubsection{Fabrication}All components can be fabricated using a commercial-level 3D printer and assembled with only a few screws within ten minutes. The compliant core is printed with thermoplastic polyurethane (TPU), and the rigid connectors are made of polylactic acid (PLA), with a total material consumption of approximately six U.S. dollars. Fabrication details can be found in the appendix on our project page.

\subsection{Deformation Sensing and Force Representation}
\label{method: Deformation Capture and Representation}
In this section, we introduce how the deformation of the compliant core is captured and transformed into force-like signals. Theoretically, we can denote that the deformation of the compliant core as $\Delta  T \in SE(3)$, which correlates with the external force $F_\text{ext}$ and torque $T_\text{ext}$ applied at the end-effector. There exists an approximately linear mapping:
\begin{equation}
\phi: \Delta  T \leftrightarrow (F_\text{ext}, \; T_\text{ext})
\end{equation}

Benefiting from the hard-soft-hard sandwich structure, the state of the compliant core is fully characterized by the relative pose between its two rigidly connected ends. 
Because the robot flange and the fiducial tag are rigidly attached to the upper and lower connectors, respectively, their poses naturally serve as reference frames to compute this transform. 

Let ${}^F T_{T,0} \in SE(3)$ denote the initial transform from the flange frame to the tag frame in the absence of external loads, and ${}^F T_{T,t} \in SE(3)$ the corresponding transform at time step $t$ under external forces and torques. 
The resulting deformation transform is then computed as:
\begin{equation}
\Delta T_t = ({}^F T_{T,0})^{-1} \, {}^F T_{T,t},
\end{equation}
where $\Delta T_t\in SE(3)$ compactly represents the relative motion of the connectors induced by external loading. Estimating the external wrench thus reduces to estimating ${}^F T_{T,t}$.

In practice, the tag pose is first obtained in the camera frame, ${}^{cam}T_T$, using OpenCV’s ChArUco marker detection. 
Because the camera is mounted on the robot flange, we use hand-eye calibration to obtain the extrinsic matrix $K_E$, which maps the camera frame to the flange frame:
\[
{}^F T_{cam} = K_E.
\]
The tag pose in the flange frame can then be expressed as:
\begin{equation}
{}^F T_{T,t} = {}^F T_{cam} \; {}^{cam} T_T.
\end{equation}

To ensure that the signal is physically meaningful, we align the translation of ${}^F T_{cam}$ with the geometric center of the lower connector and its rotation with the end-effector pose using a single rigid-body transformation from the CAD model. The resulting transform $\Delta T_t \in SE(3)$ is then decomposed into translation and Euler angles $\Delta \mathbf{T_t} \in \mathbb{R}^6$ to obtain a 6D representation directly correlated with the force-torque components. This provides an intuitive and physically consistent description of the compliant core deformation state, which is subsequently used as a force-like signal for contact-aware control and policy learning.

\subsection{Integrating Force-like Signals into Policies}
\label{method: policy}

Given the deformation signal $\Delta \mathbf{T_t}$, we aim to validate its utility across a wide range of policies for contact-rich manipulation. 
To this end, we consider two representative paradigms: classical \textbf{search-and-control policy}, which combines search strategies and control algorithms using $\Delta \mathbf{T_t}$ as both a state indicator and a reference signal. \textbf{learning-based policy}, which integrates $\Delta \mathbf{T_t}$ with other observations as an additional input to enhance policy training.

\subsubsection{Search-and-Control Policy} 
Prior works on contact-rich manipulation~\cite{chhatpar2001search,austin1997force} employ a wide range of search and control strategies, including random and spiral search trajectories, tilt-based alignment methods, and hybrid position/force controllers. 
These approaches remain widely used in real-world assembly and have proven effective in solving contact-rich tasks. 
We adopt several of these classical methods as the skill library of our search-and-control policy, denoted by 
$\pi = \{\pi_{1}, \pi_{2}, \dots, \pi_{n}\}$, 
where each $\pi_i$ represents a standard search or control skill. 
Our force-like signal $\Delta \mathbf{T_t}$ is then used either as a trigger for policy switching via simple thresholding, or as a reference for force control to regulate and adapt the robot's actions. 
Further implementation breakdown of each task can be found on \Cref{appendix:policy}.

\subsubsection{Learning-Based Policy} 
Recent advances \cite {liu2025forcemimic, xue2025reactive, hou2025adaptive} in learning-based methods, such as imitation learning, have shown strong capability for contact-rich tasks by learning from demonstrations, thereby reducing the need for manually tuning parameters and thresholds. We adopt popular imitation learning policies, such as the Action Chunking Transformer (ACT) \cite{zhao2023learning} and the Diffusion Policy (DP) \cite{chi2023diffusion}, as the backbone and train a contact-rich policy with our force-like signal $\Delta \mathbf{T_t}$ in conjunction with other observations. We then compare it against the use of a force–torque sensor or the absence of contact sensing, to further evaluate its performance in learning-based tasks.

During training the imitation policy, the policy parameters $\theta$ are optimized to minimize the discrepancy between the predicted actions $\hat{a}_t$ and the expert demonstrations $a_t$:
\begin{align}
    \hat{a}_t &= \pi_\theta(\Delta \mathbf{T_t}, I_t, \mathbf{o}_t^{\mathrm{prop}}), \\
    \theta^\star &= \arg\min_\theta \, \ell(\hat{a}_t, a_t),
\end{align}
where $\ell(\cdot)$ is a supervised imitation loss, $I_t$ is the visual image of the wrist camera, $\mathbf{o}_t^{\text{prop}}$ is the proprioception of the robotic arm, such as joint angles, and $\Delta \mathbf{T_t}$ is our force-like signal embedded by a linear layer.

Once trained, the policy uses the optimized parameters $\theta^\star$ during deployment to output actions directly:
\begin{equation}
a_t = \pi_{\theta^\star}(\Delta \mathbf{T_t}, I_t, o_t^{\text{prop}}).
\end{equation}

Further details of the deployment and performance of the learning-based policy can be found at \Cref{exp:Effectiveness in Real-World Manipulation Tasks} and \Cref{appendix:policy}.

\section{Experiment}

In this section, we perform a series of experiments aimed
at answering the following questions.
\begin{itemize}
    \item Q1: Can ShapeForce provide a high-quality force-torque representation, as well as sufficient sensitivity and stability for contact-rich manipulation? (Sec. \ref{exp:Mechanical and Material Properties})
    \item Q2: How does ShapeForce perform in real-world contact-rich tasks with different tasks and policies? (Sec. \ref{exp:Effectiveness in Real-World Manipulation Tasks})
    \item Q3: Can ShapeForce maintain long-term durability, and ensure that pretrained policies perform consistently under extensive use?  (Sec. \ref{exp: Durability and Long-Term Consistency})
\end{itemize}

\subsection{Mechanical Properties and Performance }
\label{exp:Mechanical and Material Properties}
To validate the force-torque representation quality, sensing sensitivity as well as stability of the mechanical design, we conduct the following experiments:

\subsubsection{Correlation between Wrench and Deformation}

The relationship between the wrench  (force and torque) $\mathbf{F}_t$ and the deformation-based force-like signal $\Delta \mathbf{T}_t$ generated by ShapeForce can be approximated by a linear stiffness model:
\begin{equation}
\mathbf{F}_t \approx K \, \Delta \mathbf{T}_t,
\end{equation}
where $K \in \mathbb{R}^{6 \times 6}$ is the equivalent stiffness matrix of the compliant core, and ${\mathbf{F}_t} = (F_x, F_y, F_z, T_x, T_y, T_z)^T \in \mathbb{R}^6$ is the six-dimensional ground-truth force and torque.

To validate this approximation, we collected paired data of $\mathbf{F}_t$ by a commercial force and torque sensor and force-like signal $\Delta \mathbf{T}_t$ from ShapeForce, and estimated the equivalent stiffness matrix $K$ via linear regression. We report the coefficient of determination ($R^2$) for each wrench component ($F_x$, $F_y$, $F_z$, $T_x$, $T_y$, $T_z$) in \Cref{tab:correlation}. The mean $R^2$ of 0.9577 demonstrates that the general relationship between $\Delta \mathbf{T_t}$ and the ground-truth wrench is highly linear. This near-linear correspondence suggests that our representation can serve as an effective proxy for ground-truth force-torque signals: in the context of search-and-control policies, it provides contact feedback that correspond proportionally to real interactions, enabling accurate event triggering and reference tracking; when applied within learning-based policies, it can be embedded through a simple linear layer to convey temporal contact information, making it a practical, highly cost-efficient alternative to physical force-torque sensors.
\begin{table}[H]
\centering
\caption{$R^2$ between  $\Delta \mathbf{T_t}$ and Ground-truth Wrench.}
\label{tab:correlation}
\begin{tabular}{lcccccc}
\toprule
& $F_x$ & $F_y$ & $F_z$ & $T_x$ & $T_y$ & $T_z$ \\
\midrule

$R^2$ & 0.962 & 0.9681 & 0.9587 & 0.9739 & 0.9675 & 0.9159  \\

\bottomrule
\end{tabular}

\vspace{2pt}
\end{table}

\begin{figure}[h]
    \centering
    \includegraphics[width=1\linewidth]{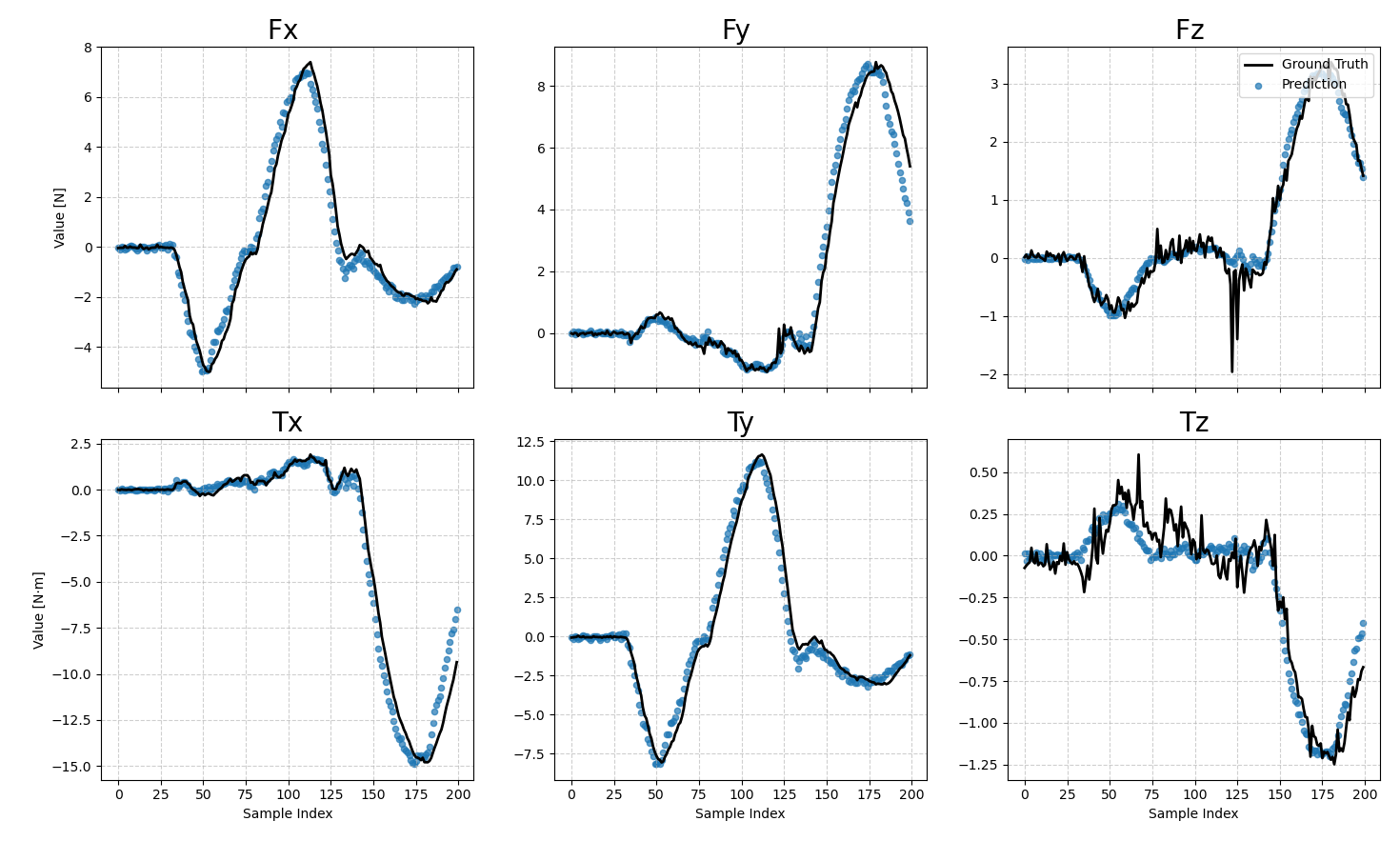}
    \caption{\textbf{Visualization of predicted vs. ground-truth signals}: the blue dots represent ShapeForce’s predictions, and the black line represents ground-truth data.}
    \label{fig:response}
\end{figure}

Moreover, if ground-truth wrench values are required, our system can reconstruct them by applying a straightforward linear transformation based on the estimated stiffness matrix $K$. A visualization example is presented in \Cref{fig:response}, demonstrating that our method can also reconstruct ground-truth force values via a simple linear transformation with high accuracy, achieving high fidelity in practice.

\vspace{2pt}
\subsubsection{Sensing Sensitivity and Stability} 

Sensitivity is a key metric for evaluating force-torque sensor performance. Since ShapeForce relies on an RGB camera for perception, its sensitivity is fundamentally limited by the visual resolution. To evaluate this for ShapeForce, we refer to the fiducial-based sensitivity analysis in prior work\cite{ouyang2020low} to calculate our theoretical sensitivity.  First, the pixel-level resolution $d_R$ is set to $1/4$ based on prior work. Using the physical tag width $w_{\mathrm{tag}}$ (mm) and its image width $w_{\mathrm{img}}$ (px), the translational sensitivities are computed as:
\[
s_z = s_x = \frac{w_{\mathrm{tag}}}{w_{\mathrm{img}}} d_R \quad (\mathrm{mm}).
\]
 And rotational sensitivity $s_{\theta y}$ is obtained from the chord-length relation: 
\[
l_{\mathrm{chord}} = 2 r \sin\left(\frac{\theta}{2}\right),
\]
where $r$ is the half-diagonal of the image patch, and set $\theta$ to be 45 degrees, and the minimum detectable angle $s_\theta$ satisfies:
\[
s_\theta = \frac{\theta}{l_{\mathrm{chord}}} d_R
\]
The axial sensitivity $s_{\theta z}$, $s_{\theta x}$, and $s_y$ are estimated via similar-triangle geometry.
Stacking the translational and rotational sensitivities into $\mathbf{s}=[s_x,s_y,s_z,s_{\theta x},s_{\theta y},s_{\theta z}]^T$, the minimum detectable wrench is estimated by:
\[
\mathbf{F}_{\min} \approx K\,\mathbf{s},
\]
where $K\in\mathbb{R}^{6\times6}$ is the calibrated stiffness matrix mentioned before. Our result shows that ShapeForce achieves a minimum detectable wrench of: 
\[
\mathbf{F}_{\min} = [\;0.41,\;0.45,\;0.87,\;0.13,\;0.12,\;0.03\;]^T~\mathrm{N~or~N\cdot m},
\]

The $\mathbf{F}_{\min}$ results showcase the sensitivity of our system to subtle forces and torques, which prove that our design is capable of detecting fine-grained contact information to support a wide range of delicate contact-rich tasks.

Also, we evaluated the stability of our mechanical design by mounting a payload on the lower connector and rotating it at different angles to evaluate the 2D displacement distance from the central axis. We mounted a payload of 0.8kg and the lever arm of a force is 6.8 cm to simulate the gravity effect of grasping an object and the end-effector. The result is shown in \Cref{tab:offset_values}. The stability results show that the soft wrist provides sufficient stiffness to support the payload and maintain accurate motion.

\begin{table}[h]
\centering
\caption{Displacement under Different Angles}
\begin{tabular}{c c c c c}
\toprule
Orientation Angle ($^\circ$) & 30 & 45 & 60 & 90 \\
\midrule
Value(mm) & 1.497 & 2.418 & 2.857 & 3.312 \\
\bottomrule
\end{tabular}
\label{tab:offset_values}
\end{table}

Overall, these results indicate that ShapeForce provides both high sensitivity, allowing the detection of subtle forces and torques, and sufficient mechanical stability to maintain precise motion under realistic payloads. Together, these properties ensure that the system delivers reliable and physically meaningful feedback for a variety of contact-rich tasks.

\subsection{Effectiveness in Real-World Manipulation Tasks}
\label{exp:Effectiveness in Real-World Manipulation Tasks}

To directly validate ShapeForce's task performance, we evaluate performance across six representative contact-rich manipulation tasks with the aforementioned two types of policies: search-and-control policy and learning-based policy. The details of the implementation and results are as follows: 

\begin{figure}[t]
    \centering
    \includegraphics[width=0.9\linewidth]{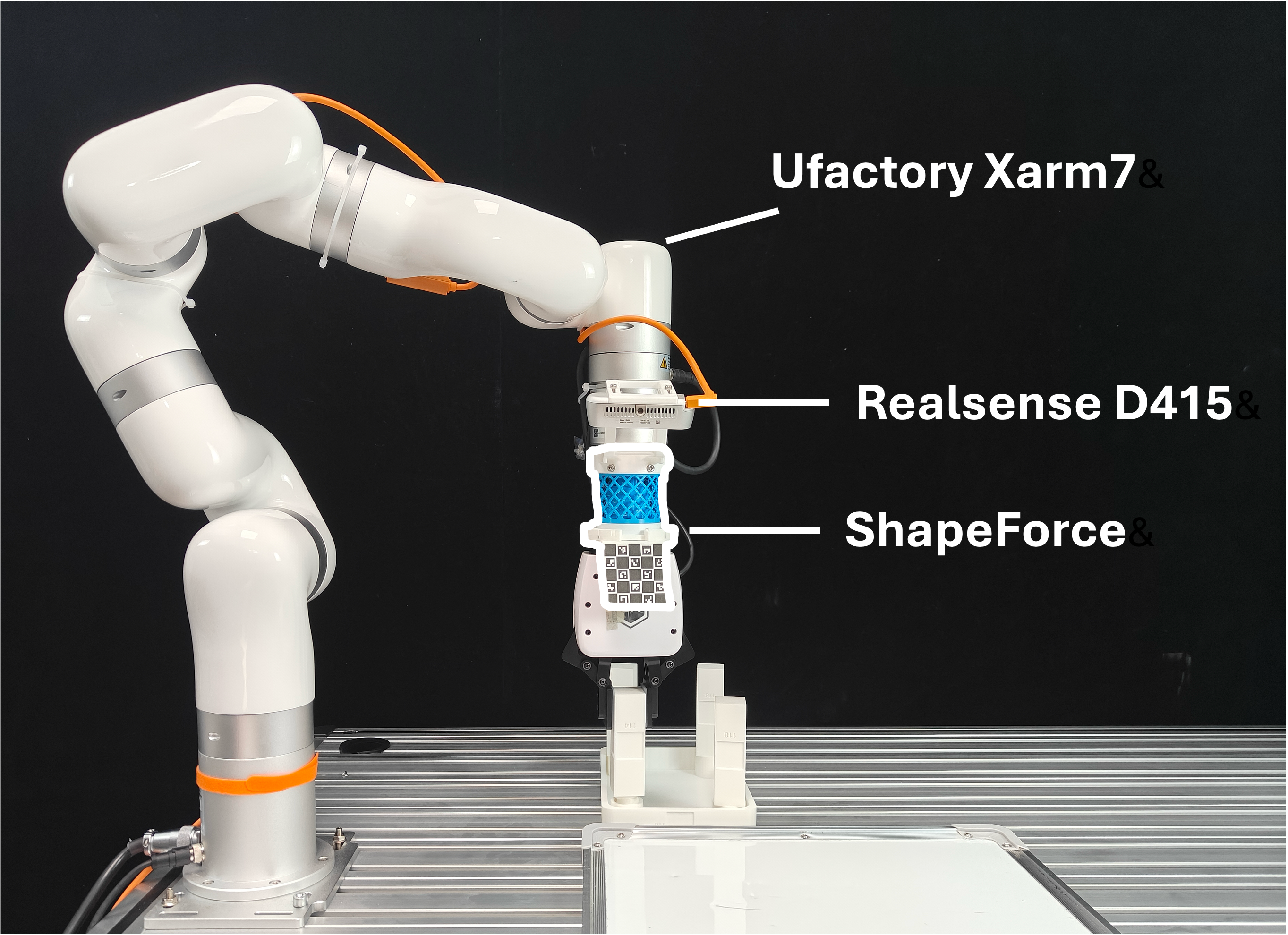}
    \caption{\textbf{Robot Experiment Setup}: We use Ufactory xArm7 robotic arm for manipulation and an Intel RealSense D435 for visual perception. ShapeForce is mounted on the robotic wrist to obtain the contact information.}
    \label{fig:setup}
\end{figure}

\subsubsection{Search-and-Control Policy}

\begin{table}[t]
\centering
\caption{Task Success Rate of Search-and-Control Policy}
\label{tab:task_success_rate}
\begin{tabular}{lcc}
\toprule
\textbf{Task} & \textbf{With F/T Sensor} & \textbf{ShapeForce} \\
\midrule
Peg Insertion        & 100\% & 100\% \\
USB Insertion      & 100\% & 90\% \\
Toy Desk Assembly          & 65\% &  80\% \\
Bottom Cap Screw Tightening  & 100\% & 100\% \\
Whiteboard Wiping           & 100\% & 100\% \\
Maze Exploration           & 100\% & 100\% \\
\bottomrule
\end{tabular}
\end{table}
In this setting, we use ShapeForce as a six-axis force–torque sensor to provide contact feedback for search-and-control policies. For insertion tasks, we employ search-based algorithms to locate the target, while classical controllers such as PID are used to track desired forces during tasks like whiteboard wiping. Other state transitions are triggered by simple threshold-based events.

\begin{figure}[t]
    \centering
    \includegraphics[width=1\linewidth]{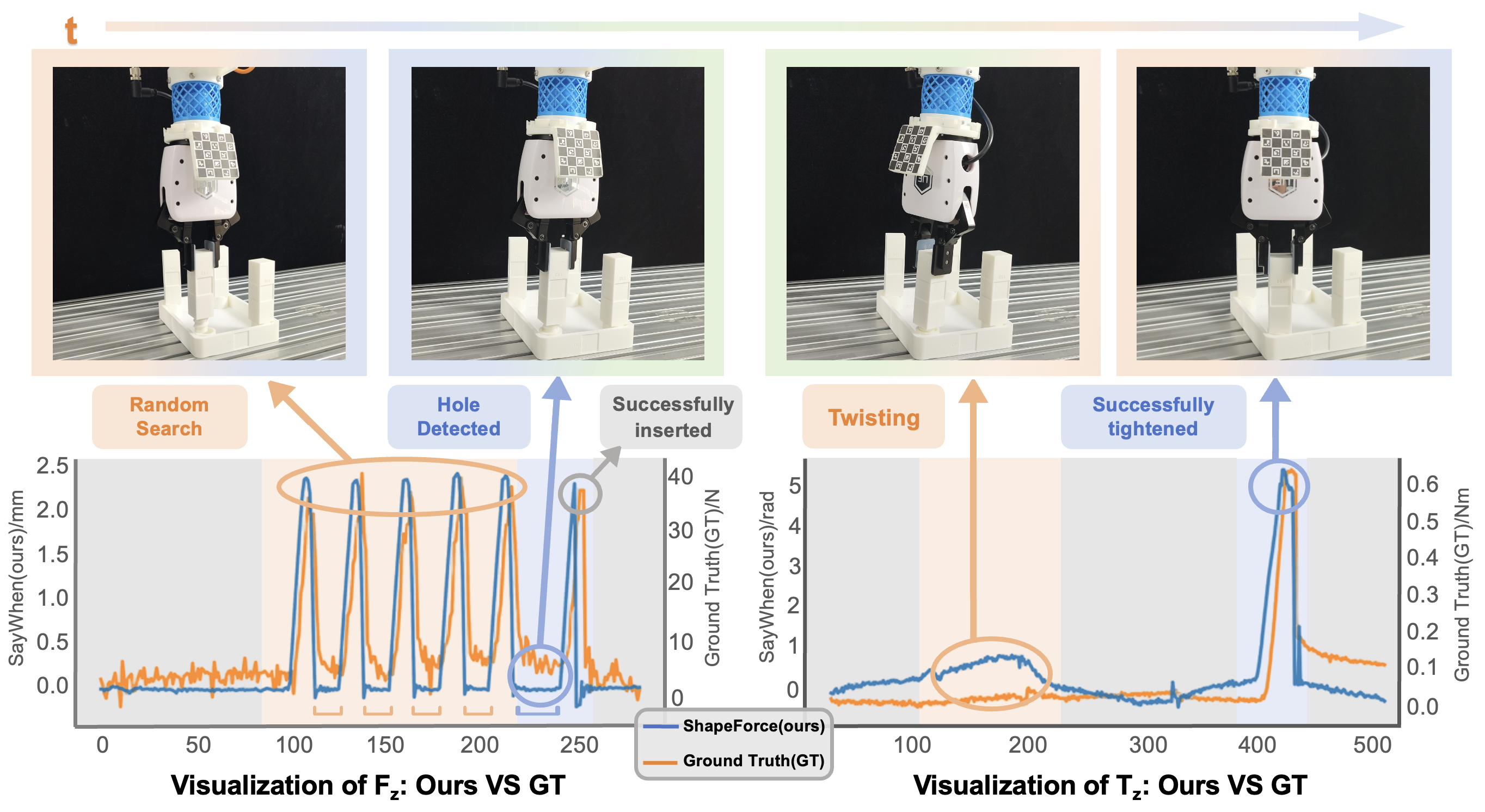}
    \caption{\textbf{Visualization of Toy Desk Assembly}: Comparison of the $F_z$ and $T_z$ components of our force-like signal with ground-truth measurements. After proportional scaling, the signals exhibit close alignment, confirming that our representation preserves the essential contact information.}
    \label{fig:furniturebench}
\end{figure}

For each task, we evaluate the performance of ShapeForce and a 6-axis force-torque sensor from UFactory, which provides ground-truth measurements. Performance is measured by success rate, with outcomes assessed by a human expert due to the explicit nature of the tasks. We do not compare with a setting without any contact feedback, as search-and-control policies cannot be designed without it, and the success rate ought to be zero. We evaluate each both setup in task for 20 times to ensure statistical reliability.

Results are summarized in \Cref{tab:task_success_rate}. Our system achieves a success rate comparable to that of a 6-axis force-torque sensor, demonstrating that the proposed force-like signal can serve as an effective substitute for ground-truth measurements. Unlike commercial sensors, however, our approach requires no expensive hardware and is seamlessly integrated into the robot, offering a low-cost yet highly reliable alternative for search-and-control manipulation policies.

The only noticeable drop occurs in the USB insertion task, where the inherent compliance of our design introduces slight uncertainty. Despite our efforts to mitigate this effect, tasks that demand extreme precision in contact detection, such as USB insertion, remain more challenging. In particular, distinguishing between the correct and incorrect orientations during insertion requires detecting differences within a tolerance of about 1 mm, which can be easily falsely triggered by noise or compliance-related uncertainty.

However, our system outperforms the 6-axis force-torque sensor in the toy desk assembly task, which requires highly precise alignment of the insertion direction to properly engage the threads. Here, the compliant nature of our design becomes advantageous: even with slight orientation errors, the passive deformation of the wrist compensates for misalignment, allowing smooth insertion and tightening. In contrast, a rigid wrist would cause large contact forces under the same condition, often leading to slippage between the leg and the gripper and ultimately task failure. This highlights the inherent robustness provided by passive compliance. Overall, ShapeForce performs a comparable performance with force-torque sensor in classical search-and-control policy.

\subsubsection{Learning-based Policy}

To validate that our force-like signal $\Delta \mathbf{T}_t$  can be applied in learning-based policy therefore further prove that it can be a effective alternative of ground-truth force-torque signals, we adopt two imitation learning policies Action Chunking Transformer (ACT) \cite{zhao2023learning} and the Diffusion Policy (DP) \cite{chi2023diffusion} as our backbones to evaluate. 

We first construct a 15-step temporal history of $\Delta \mathbf{T}_t$, encode it via a linear projection layer, and concatenate it with other observations to train the policies. This design enhances the contact-awareness of both policies. Expert demonstrations are collected using the Gello teleoperation system \cite{wu2024gello}, with 50 trajectories per task for each policy. To improve robustness and prevent overfitting, we add small random positional and orientational variation to the objects during both data collection and evaluation. We benchmark our method on four contact-rich manipulation tasks: peg insertion, USB insertion, whiteboard wiping, and toy desk assembly. Performance is compared against two baselines: (i) policies trained without any force-related input, and (ii) policies using ground-truth force-torque signals.

The results of our experiments are summarized in \Cref{tab:imitation_results}. Across all tasks, policies trained without any force or deformation signal perform significantly worse. These policies can only replicate demonstrated motion patterns without perceiving contact states, which causes them to learn an ``average'' behavior that fails under dynamic variations. This limitation is particularly evident in tasks requiring sustained contact, such as whiteboard wiping. In this task, visual occlusion from the eraser often prevents the policy from accurately estimating the surface distance by image. Meanwhile, the board’s orientation and height distance vary across trials. Successfully wiping requires consistent contact pressure. Without force-related signals, success rates for both ACT and DP policies drop below 30\%. In contrast, when incorporating either ground-truth force or our deformation signal $\Delta \mathbf{T}_t$, both policies exceed 90\% success rate. This huge leverage demonstrate the critical role of force-related observations for contact-rich imitation learning.

Comparing ShapeForce with the ground-truth sensor, our deformation signal $\Delta \mathbf{T}_t$ achieves success rates comparable to those of the force sensor, further demonstrating the linear relationship between $\Delta \mathbf{T}_t$ and the wrench, which is approximately equivalent for the learning-based method. And in some cases, like the toy desk assembly task, ShapeForce enables the ACT policy to reach 80\% success, compared to 65\% with the force sensor, highlighting the advantages of passive compliance. Overall, these results show that ShapeForce generalizes well in imitation learning and serves as an effective, low-cost alternative to force–torque sensors.

\begin{table}[t]
\centering
\caption{Task Success Rate of Learning-based Policy}
\label{tab:imitation_results}
\resizebox{\columnwidth}{!}{%
\begin{tabular}{lccc}
\toprule
\textbf{Task} & \textbf{Without Sensor} & \textbf{With F/T Sensor} & \textbf{ShapeForce} \\
\midrule
Peg Insertion (ACT) & 60\% & 85\% & 90\% \\
Peg Insertion (DP) & 65\% & 75\% & 85\% \\
USB Insertion (ACT) & 40\% & 85\% & 75\% \\
USB Insertion (DP) & 35\% & 80\% & 75\% \\
Whiteboard Wiping (ACT) & 25\% & 95\% & 95\% \\
Whiteboard Wiping (DP) & 30\% & 90\% & 90\% \\
Toy Desk Assembly (ACT) & 45\% & 65\% & 80\% \\
Toy Desk Assembly (DP) & 40\% & 50\% & 70\% \\

\bottomrule
\end{tabular}%
}
\end{table}

\subsection{Durability and Sensing Consistency }
\label{exp: Durability and Long-Term Consistency}

\begin{table}[t]
\centering
\caption{Task Success Rate after Aging Test}
\label{tab:aging test}
\resizebox{\columnwidth}{!}{%
\begin{tabular}{lcccc}
\toprule
\textbf{Task} & \textbf{1k} & \textbf{5k} & \textbf{10k} & \textbf{20k} \\
\midrule
Whiteboard Wiping (S\&C) & 100\% & 100\% & 100\% & 95\% \\
Toy Desk Assembly (S\&C) & 80\% & 80\% & 80\% & 80\% \\
Peg Insertion (ACT) & 85\% & 85\% & 85\% & 85\% \\
Toy Desk Assembly (ACT) & 80\% & 80\% & 60\% & 60\% \\
Toy Desk Assembly (ACT-finetune) & 85\% & 80\% & 80\% & 80\% \\
\bottomrule
\end{tabular}%
}
\end{table}

As a low-cost sensing system made from 3D-printed compliant materials, durability is crucial for practical use, so we performed an aging test to evaluate the durability of the sensor. We fix the lower connector to a fixture and move the robotic arm to apply force in various directions to the compliant core. Generally, aging of the elastic material can cause some plastic deformation and degradation of its linear response characteristics. To comprehensively evaluate the influence, we adopt parameters-tuned search-and-control policies and a pretrained ACT model for multiple tasks after aging 1k, 5k, 10k, and 20k times. 

Results are shown in \Cref{tab:aging test}. For search-and-control policies, performance was evaluated on whiteboard wiping and toy desk assembly. Stiffness degradation lowers the force threshold, reducing the applied force under the same conditions. This causes a minor drop in whiteboard wiping performance after 20k repetitions, but the magnitude remains small, demonstrating the robustness and consistency of our system under search-and-control policy. For learning-based policy, we tested the pretrained ACT model directly on peg insertion and toy desk assembly. The result demonstrates that ACT can perform consistently in relatively simple tasks like peg insertion, but in complicated tasks like toy desk assembly, our pretrained policy fails due to the degradation of linearity and elastic property. However, after fine-tuning with five demonstrations, the performance can quickly recover to a success rate nearly matching that of a brand-new sensor. In a nutshell, the aging test demonstrates that ShapeForce has high long-term durability and policy consistency under extensive use, and can be rapidly restored through minimal fine-tuning.

\section{conclusion and discussion}
In this paper, we introduce ShapeForce, a low-cost, plug-and-play soft robotic wrist that provides force-like signals enabling contact-rich tasks. We proved the efficiency of this force-like signal through extensive experiments on various contact-rich tasks with different policies, where ShapeForce significantly improves success rates and is comparable to commercial force-torque sensors. This work not only offers a cost-efficient, easily reproducible alternative to traditional force-torque sensors but also provides a fundamental insight: the trade-off between accuracy and simplicity in sensing systems can be effectively optimized by leveraging task-relevant features to construct an equivalent representation.

However, ShapeForce has some limitations. Its passive compliance makes precise motion difficult under high payloads, and its material properties are not fully optimized. These limitations stem from its original design goals of being low-cost, accessible, and reproducible. Despite these limitations, we hope this work will encourage and inspire more researchers to study further and explore the contact-rich tasks.

\section{Appendix}
\subsection{Fabrication Details}
Directly inferring material parameters from desired mechanical properties is challenging due to the heterogeneous, multi-component nature of the structure. To address this, we conducted extensive empirical tests across representative tasks, systematically varying material and structural parameters. Based on these experiments, we selected a set of parameters that best balance all design objectives and used them for subsequent deployment.

To contextualize this parameter selection, we revisit our overarching design requirements. For the upper layer, we aim to (i) enhance sensitivity along the Z-axis while maintaining sufficient structural stiffness under lateral impact, and (ii) minimize displacement under large lateral loads. To achieve these performance targets, we systematically explored key 3D printing parameters that influence mechanical strength, including infill density (15–80\%), number of wall loops (2–6), and dimensional parameters of the compliant core geometry. As for the lower layer, since it is mechanically decoupled from the upper layer, we predetermined its fabrication parameters. Instead, after finalizing the upper-layer parameters, we fine-tuned the thickness of the connecting structures between the inner and outer rings of the lower layer based on its actual mechanical response.

After an iterative experimental process, we converged on a parameter set that effectively balances the competing requirements of sensitivity and stiffness, as detailed in Table~\ref{table:structural_parameters}.
\begin{table}[h]
\centering
\caption{Key Materiel and Structural parameters.}
\begin{tabular}{l l}
\toprule
\textbf{Type} & \textbf{Detail} \\
\midrule
Bambu Lab TPU 95A (core)   & infill 80\% , wall loops 2 \\
Bambu Lab PLA Basic (frame)   & infill 30\% , wall loops 2 \\
Compliant Core  & thickness of beams 2.3 mm \\
Compliant Core  & length of beams 7.3 mm \\ 
Compliant Core  & width of beams 1.5 mm \\ 
Compliant Core   & thickness of sleeve 6 mm \\
Wave-like Support   & thickness 3 mm \\
Layer height   & 0.2 mm \\

\bottomrule
\label{table:structural_parameters}
\end{tabular}
\end{table}

\subsection{Real-World Experiment Details}
\label{appendix:policy}
\subsubsection{Search-and-Control Policy}

To address the challenges of contact-rich manipulation, we adopt a hybrid framework that combines exploratory motion strategies with reactive force control, which is referred to as a search-and-control policy. This paradigm integrates classical algorithms for tasks involving uncertain or unmodeled contacts: search routines are used to resolve geometric uncertainty, while force control strategies (e.g., PID with force feedback) regulate interaction forces during insertion or wiping. The switch of policies are decided by preset thresholds. By combining perception-driven search with low-level force control, the policy enables robust execution of contact-rich tasks.

We evaluate this approach on five representative real-world tasks: peg insertion, USB insertion, toy desk assembly, bottom-cap screw tightening, and whiteboard wiping. To address this series of contact-rich tasks, we adopt representative baseline algorithms, including classical search and control methods. A detailed breakdown about policies of each task is provided below.

\begin{itemize}
\item \textbf{\textit{Peg Insertion:}} (i) a visual servo search to approach the hole, with threshold $\tau_1$ triggered when the end-effector reaches the search surface; (ii) a directional surface search, where threshold $\tau_2$ is triggered upon detecting the hole; (iii) a compliance-based insertion, continuing until threshold $\tau_3$ is reached. 

\item \textbf{\textit{USB Insertion:}} Similar to peg insertion but with orientation ambiguity: the plug can align in two ways, yet only one allows full insertion. The policy detects the hole via $\tau_1$, attempts insertion, and if $\tau_2$ signals failure, rotates 180° to retry.

\item \textbf{\textit{Bottom-cap Screw Tightening:}} (i) a visual servo search to approach and align until $\tau_1$ triggered; (ii) a force-guided screw motion until threshold $\tau_2$ is reached.
\item \textbf{\textit{Toy Desk Assembly:}} Combines first and second type, inserting and screwing the tightening of table legs.  
\item \textbf{\textit{Whiteboard Wiping:}} (i) visual servoing to approach the surface until $\tau_1$ is triggered; (ii) using PID maintains consistent contact force by chasing $\tau_2$ to wiping.  
\item \textbf{\textit{Maze Exploration:}} (i) heuristic direction-prioritized greedy search until collision force $\tau_1$ is triggered; (ii) real-time force-monitored reverse retreat until safe force $\tau_2$ is reached, registering a transient escape node; (iii) direction-level dead-end learning with backtracking and path pruning.
\end{itemize}

\subsubsection{Learning-Based Policy}
\begin{figure}[h]
    \centering
    \includegraphics[width=0.9\linewidth]{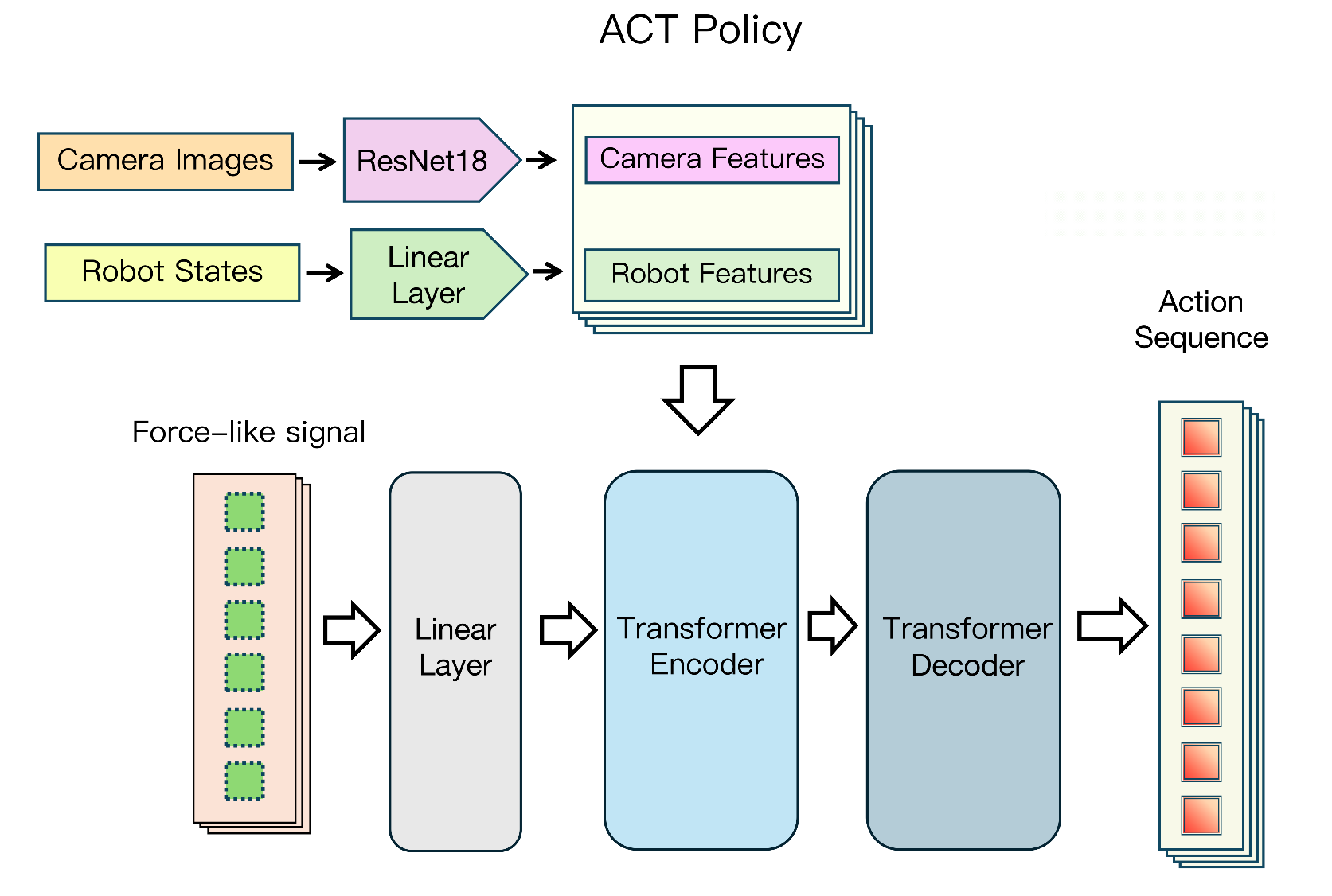}
    \caption{Structure of the ACT-based Network.}
    \label{fig:act}
\end{figure}

\begin{figure}[h]
    \centering
    \includegraphics[width=0.9\linewidth]{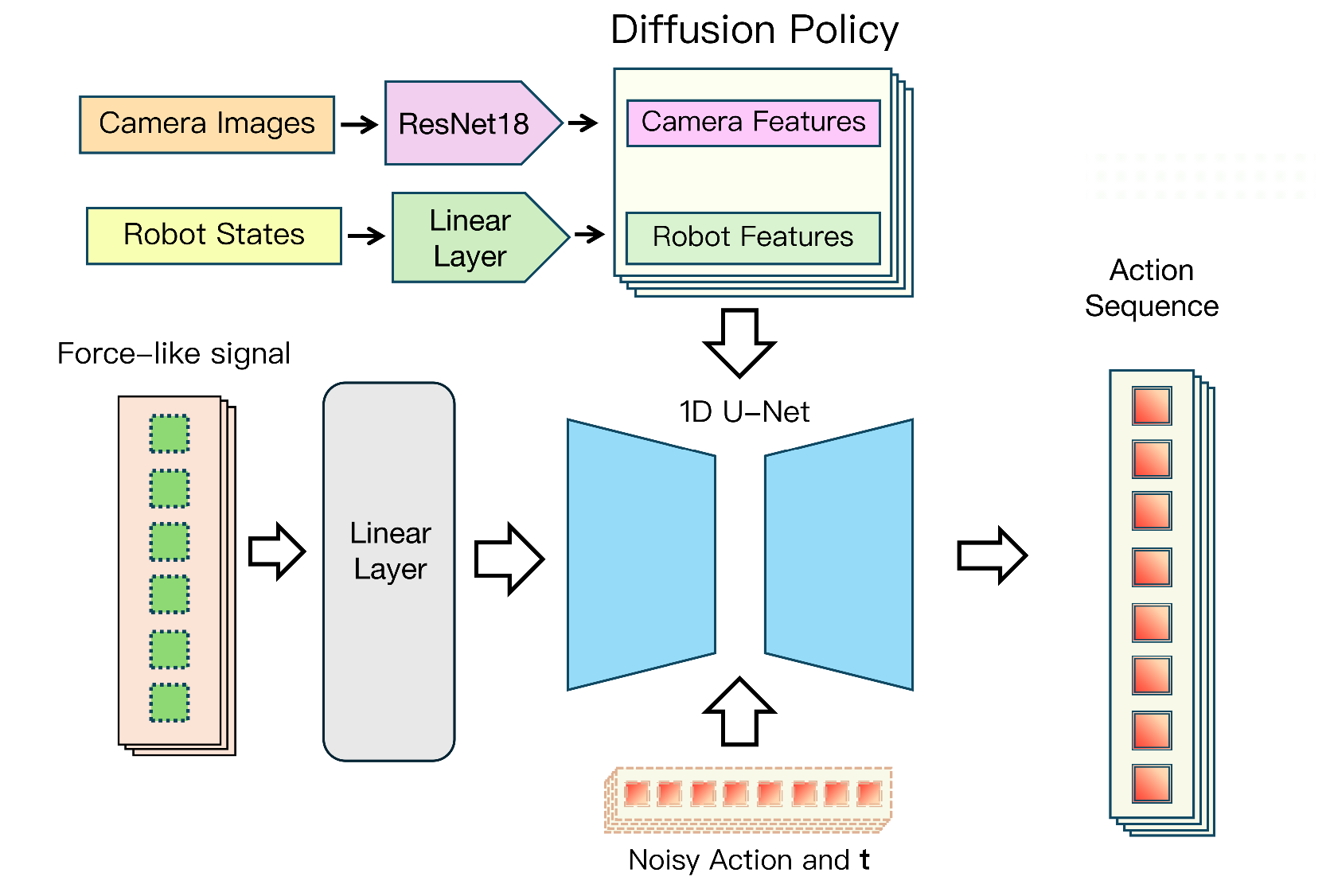}
    \caption{Structure of the DP-based Network.}
    \label{fig:dp}
\end{figure}

We integrate our force-like signal into both the ACT (Action Chunking Transformers) in \Cref{fig:act} and DP (Diffusion Policy) in \Cref{fig:dp}. The imitation learning data is collected using Gello hardware. For the force-like signal, we apply polynomial filtering as preprocessing during data collection and also perform real-time polynomial filtering during deployment. When used in the policy, the force-like signal is stacked over the past 15 timesteps and then fed into the policy through a linear layer.
For the ACT policy, cond vectors compose of visual features (from frozen ResNet-18), robot joint states, and processed force-like history, are fed into a transformer encoder-decoder with 4 layers each. The model predicts a chunk of 45 future actions in a single forward pass. For the Diffusion Policy, the similar conditioning inputs are fed into a 1D U-Net Network to guide a denoising process over 48 action steps.
\begin{table}[h]
\centering
\caption{Comparison of key hyperparameters for ACT and Diffusion Policy (DP).}
\begin{tabular}{l l l}
\toprule
\textbf{Parameter} & \textbf{ACT} & \textbf{Diffusion Policy (DP)} \\
\midrule
Chunk size  & 45 & 48 \\
Hidden dim & 256 & 256 \\
Batch size & 64 & 64 \\
Learning rate & $1 \times 10^{-4}$ & $1 \times 10^{-4}$ \\
Weight decay & $1 \times 10^{-3}$ & $1 \times 10^{-3}$ \\
Transformer layers & 4 / 4 & — \\
Diffusion time steps & — & 100 \\
Vision backbone & ResNet-18 & ResNet-18\\
\bottomrule
\label{table:policy_hyperparameters}
\end{tabular}
\end{table}

\subsection{Adaptability}
\begin{figure}[h]
    \centering
    \includegraphics[width=0.9\linewidth]{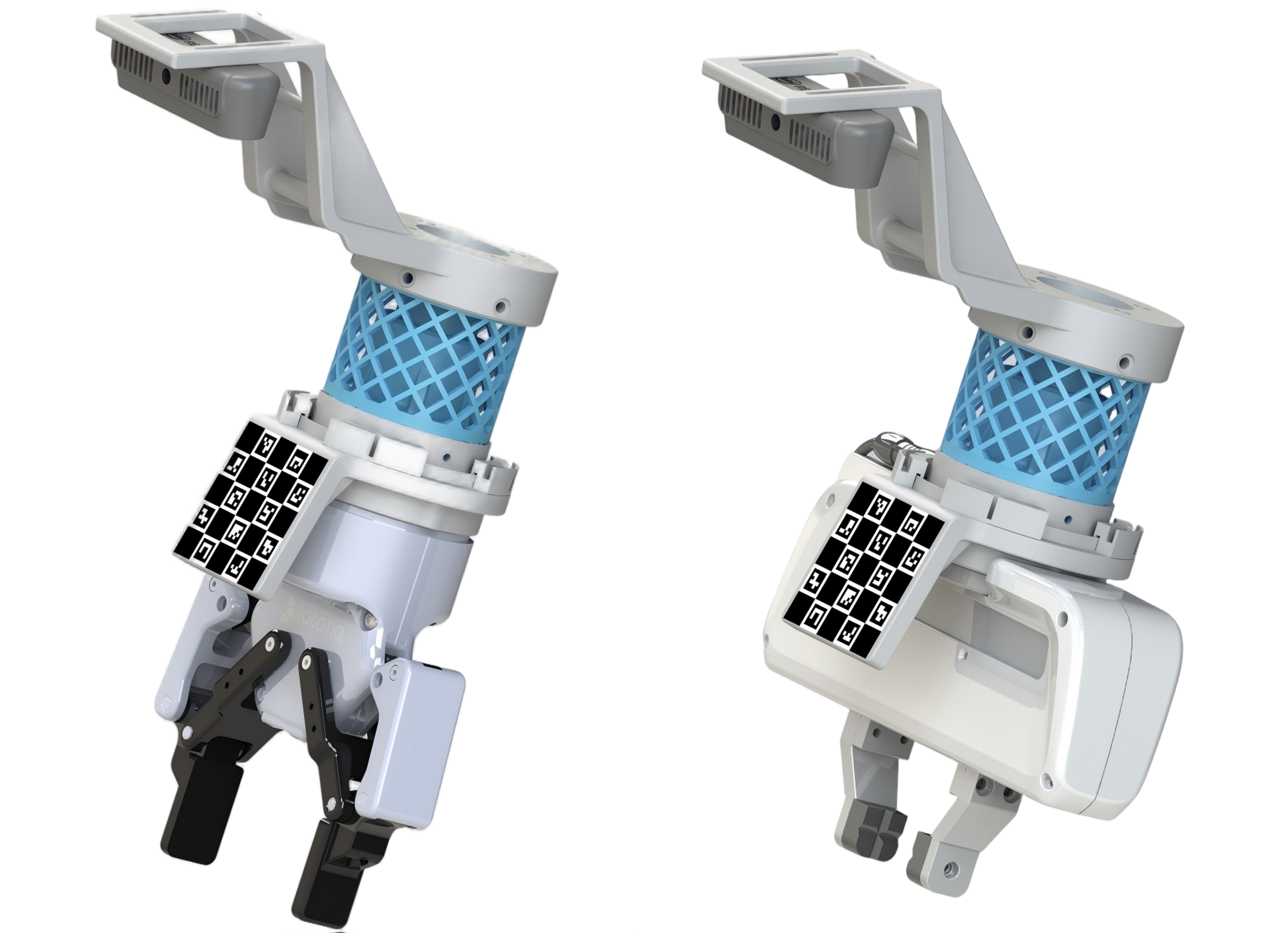}
    \caption{Examples of our adapter design installed on two different grippers, demonstrating the flexibility to accommodate diverse hardware platforms.}
    \label{fig:gripper}
\end{figure}
To ensure broad applicability across different robotic platforms, we designed a set of lightweight and modular adapters as shown in \Cref{fig:gripper}. These adapters allow our system to be seamlessly mounted on various types of grippers and robotic arms, rather than being restricted to a single hardware configuration. With this modular design, users can quickly switch between end-effectors or deploy the system on different manipulators without additional customization, significantly enhancing its efficiency and flexibility for real-world deployment in novel robots.

\subsection{Exploratory Designs}
\begin{figure}[h]
    \includegraphics[width=0.9\linewidth]{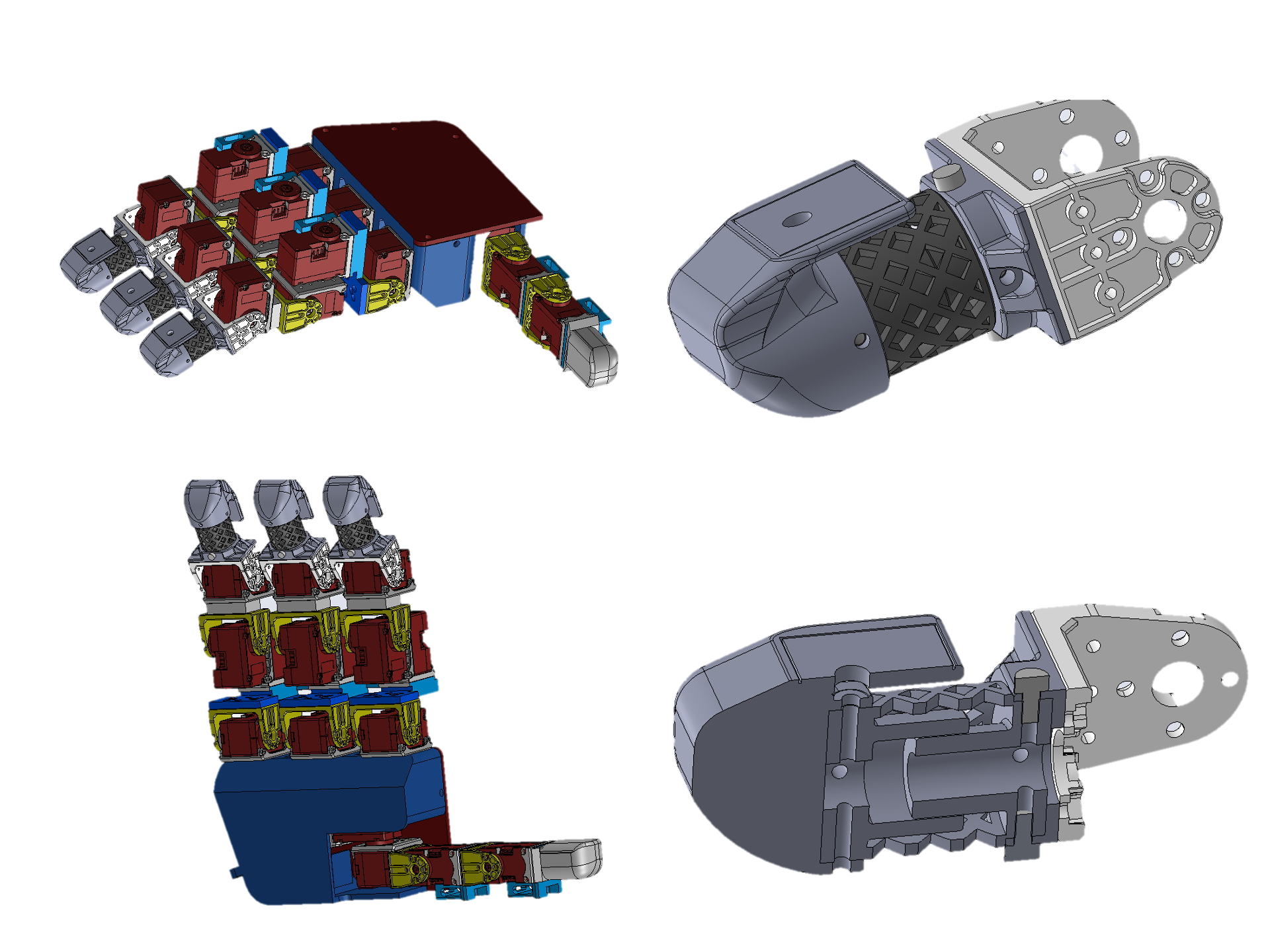}
    \caption{An exploratory example where ShapeForce is adapted from a wrist-mounted configuration to the fingertips of a dexterous hand, showcasing its potential generalizability across different end-effectors.}
    \label{fig:hand}
\end{figure}
The ShapeForce structure exhibits strong generality and is not limited to a single type of robotic manipulator. In addition to its wrist-mounted configuration, we explored mounting ShapeForce at the fingertips of a dexterous hand, which is shown in \Cref{fig:hand}. In this setup, the pose of one end can be obtained through forward kinematics (FK), while the other end can be estimated using a fiducial tag. Together, these allow accurate deformation estimation and thus reliable force prediction, even in this more complex configuration. This exploratory design highlights the potential of ShapeForce to be generalized across diverse end-effectors. We envision this direction as a promising step toward making ShapeForce a universal and adaptable sensing module that can be seamlessly integrated into the end-effectors of various robotic platforms.

\clearpage



\bibliographystyle{IEEEtran}
\bibliography{references}




\end{document}